\documentclass{article}

\usepackage{arXiv}
\usepackage[utf8]{inputenc}
\usepackage[T1]{fontenc}
\usepackage{pifont}
\usepackage{graphicx}
\usepackage{url}
\usepackage{booktabs}
\usepackage{amsfonts}
\usepackage{microtype}
\usepackage{multirow}
\usepackage{natbib}
\usepackage{bbm}
\setcitestyle{numbers,square}
\usepackage{amsmath}
\usepackage{caption}
\usepackage{hyperref}
\usepackage{xcolor}
\definecolor{mydarkblue}{rgb}{0, 0.08, 0.45}
\hypersetup{
    colorlinks=true,
    linkcolor=mydarkblue,
    filecolor=mydarkblue,
    urlcolor=mydarkblue,
    citecolor=mydarkblue,
    pdftitle={When to Ponder: Adaptive Compute Allocation for Code Generation via Test-Time Training},
}
\usepackage{tikz}
\usetikzlibrary{shapes.geometric, arrows.meta, positioning, calc}

\newcommand{\cmark}{\ding{51}}
\newcommand{\xmark}{\ding{55}}

\title{When to Ponder: Adaptive Compute Allocation for Code Generation via Test-Time Training}
\renewcommand{\shorttitle}{When to Ponder}

\date{December 31, 2025}

\author{
  \href{https://orcid.org/0000-0001-7372-9423}{\includegraphics[scale=0.06]{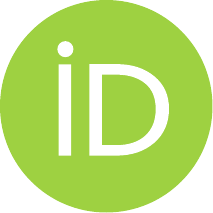}\hspace{1mm}Gihyeon Sim} \\
{\normalfont Dongpae High School, Paju, Gyeonggi-do, Republic of Korea} \\
  \texttt{world@worldsw.dev} \\
}

\begin{document}

\maketitle

\begin{abstract}
Large language models apply uniform computation to all inputs, regardless of difficulty. We propose \emph{PonderTTT}, a gating strategy using the TTT layer's self-supervised reconstruction loss to selectively trigger Test-Time Training (TTT) updates. The gating decision itself is \textbf{training-free}---requiring no learned classifier or auxiliary networks; only a single scalar threshold is \textbf{initially calibrated} on unlabeled data and \textbf{continuously adapted} via EMA to maintain target update rates. Our experiments with GPT-2 models (124M to 1.5B) on code language modeling (The Stack v2, teacher-forced perplexity) demonstrate that this signal is \textbf{inference-compatible}, requiring no ground-truth labels. Our Reconstruction Gating achieves \textbf{82--89\% Oracle Recovery} while being fully training-free, significantly outperforming Random Skip baselines (up to 16\% lower loss on OOD languages).
\end{abstract}

\keywords{Test-Time Training \and Adaptive Computation \and Language Models \and Code Generation \and Sample Efficiency \and Dynamic Inference}

\section{Introduction}
\label{sec:intro}

Standard Transformer models operate on a fixed computational graph: every token processes the same number of layers and attention heads. While effective, this rigidity creates inefficiency. Consider code generation: producing a standard \texttt{import} statement requires far less computation than implementing a dynamic programming algorithm. A fixed-compute model must either be over-provisioned for simple cases or under-provisioned for complex ones.

Prior approaches to adaptive computation, such as Mixture-of-Experts (MoE) or Early Exit strategies, focus on routing tokens or skipping layers but do not modify the model's representations based on input context. Test-Time Training (TTT) offers an alternative: the model's parameters are updated during inference to adapt to the current input. However, standard TTT applies updates uniformly (e.g., gradient descent on every token), reintroducing computational inefficiency.

We propose \emph{PonderTTT}, which focuses on finding the optimal ``When to Update'' signal. We define ``Pondering'' in this context as the deliberate, adaptive allocation of the update budget---deciding \emph{whether} to learn from the current context rather than \emph{how long} to think. Unlike PonderNet's geometric halting mechanism for layer-wise computation, our approach controls \emph{parameter updates} at inference time. We empirically observe that the TTT layer's self-supervised reconstruction loss provides a training-free \textbf{Reconstruction Gating} strategy that works efficiently and robustly.

\begin{itemize}
    \item We demonstrate that \textbf{TTT Reconstruction Loss} is an inference-compatible proxy for learning potential. This self-supervised signal is available during inference without ground-truth labels.
    \item We introduce ``Reconstruction Gating,'' a threshold-based gating strategy with EMA adjustment, and analyze its effectiveness across model scales.
    \item We provide empirical analysis showing that Reconstruction Gating achieves \textbf{82--89\% Oracle Recovery} while being fully training-free, significantly outperforming Random Skip baselines.
\end{itemize}

\section{Related Work}

\textbf{Adaptive Computation.} Efforts to move beyond the fixed-compute paradigm include Universal Transformers \cite{dehghani2019universal}, which loop over layers dynamically, and Early Exit models \cite{schuster2022confident}, which produce predictions at intermediate layers. PonderNet \cite{banino2021pondernet} introduced a probabilistic halting mechanism trained via variational inference. Unlike these architectural modifications, our work focuses on adapting the \textit{parameters} of the model (fast weights) dynamically.

\textbf{Test-Time Training (TTT).} TTT \cite{sun2020test} was originally proposed for generalization in vision tasks. Recently, TTT-LM \cite{sun2024learning} adapted this to language modeling by augmenting the transformer architecture with a self-supervised adaptation layer that learns from historical context, proposing both TTT-Linear and TTT-MLP variants. We adopt TTT-Linear for computational efficiency, as it requires only matrix-vector operations rather than the additional nonlinearities in TTT-MLP. Our work builds directly on this layer but addresses the open problem of \textit{when} to trigger these updates.

\textbf{Meta-Learning.} Our approach can be viewed as ``learning to learn,'' or meta-learning \cite{finn2017model}. The static weights of our model serve as meta-parameters that determine how the fast weights should change. We extend this by \textit{deriving} an input-conditioned update schedule from the TTT layer's learned reconstruction signal---the schedule itself requires no additional training.

\section{Method}
\label{sec:method}

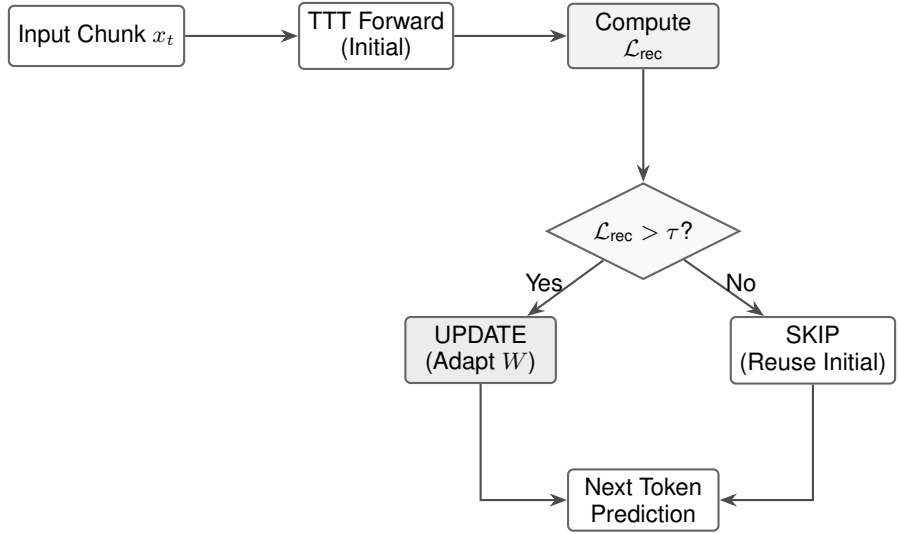
\begin{figure}[t]
\centering
\begin{tikzpicture}[
    node distance=1.5cm,
    >=Stealth,
    font=\small\sffamily,
    box/.style={rectangle, draw=black!60, thick, rounded corners=2pt, minimum width=2.0cm, minimum height=0.8cm, align=center, fill=white},
    decision/.style={diamond, draw=black!60, thick, aspect=2, minimum width=1.5cm, align=center, fill=gray!5},
    arrow/.style={->, thick, draw=black!70}
]
    
    \node[box] (input) {Input Chunk $x_t$};
    \node[box, right=of input] (forward) {TTT Forward\\(Initial)};
    \node[box, fill=gray!10, right=of forward] (loss) {Compute\\$\mathcal{L}_{\text{rec}}$};
    \node[decision, below=of loss] (threshold) {$\mathcal{L}_{\text{rec}} > \tau$?};
    \node[box, fill=gray!15, below left=0.8cm and 0.5cm of threshold] (update) {UPDATE\\(Adapt $W$)};
    \node[box, below right=0.8cm and 0.5cm of threshold] (skip) {SKIP\\(Reuse Initial)};
    \node[box, below=2.5cm of threshold] (output) {Next Token\\Prediction};
    
    \draw[arrow] (input) -- (forward);
    \draw[arrow] (forward) -- (loss);
    \draw[arrow] (loss) -- (threshold);
    \draw[arrow] (threshold) -- node[left, pos=0.4, font=\footnotesize\sffamily] {Yes} (update);
    \draw[arrow] (threshold) -- node[right, pos=0.4, font=\footnotesize\sffamily] {No} (skip);
    \draw[arrow] (update) |- (output);
    \draw[arrow] (skip) |- (output);
    
\end{tikzpicture}
\caption{\textbf{PonderTTT Architecture.} Each chunk undergoes an initial TTT forward pass. The reconstruction loss $\mathcal{L}_{\text{rec}}$ determines whether to UPDATE (adapt weights and re-forward) or SKIP (reuse initial forward result). The gating decision requires no learned classifier.}
\label{fig:architecture}
\end{figure}

We consider a causal language modeling task where the input sequence $X = (x_1, \dots, x_T)$ is processed in chunks $C_1, \dots, C_K$. The model parameters consist of slow weights $\theta_{\text{slow}}$ (frozen backbone) and fast weights $\theta_{\text{fast}}$ (TTT layer, denoted $W_t$ below). Figure~\ref{fig:architecture} illustrates our gating mechanism.

\subsection{Preliminaries: TTT-Linear Update}
Following \cite{sun2024learning}, the TTT layer maintains a hidden state $W_t$ (fast weight) which is updated via a self-supervised reconstruction task. For an input chunk $x_t$, the update rule is:
\begin{equation*}
    W_{t+1} = W_t - \eta \nabla_{W_t} \mathcal{L}_{\text{rec}}(W_t; x_t)
\end{equation*}
where $\eta$ is a position-dependent learnable learning rate. The self-supervised reconstruction loss $\mathcal{L}_{\text{rec}}$ reconstructs the residual $(V - K)$ from $K$:
\begin{equation*}
    \mathcal{L}_{\text{rec}}(W_t; x_t) = \| \text{LayerNorm}(K \cdot W_t + b_t) - (V - K) \|^2
\end{equation*}
The output then adds $K$ back via residual connection, effectively reconstructing $V$. Here, $K, V \in \mathbb{R}^{B \times T \times d}$ are projections from the current chunk's hidden states. Following TTT-Linear~\cite{sun2024learning}, $K$ and the test-view $Q$ share a base projection (\texttt{wq} in code), differentiated by separate causal convolutions; $V$ uses an independent projection (\texttt{wv}).

\textbf{Training Objective.} The TTT layer is trained by minimizing the combined loss:
\begin{equation*}
    \mathcal{L}_{\text{total}} = \mathcal{L}_{\text{CE}} + \beta \mathcal{L}_{\text{rec}}
\end{equation*}
where $\beta=0.1$ (see Appendix Table~\ref{tab:hyperparams} for full configuration).

\textbf{Causal Masking.} The TTT update uses a lower-triangular attention mask to ensure causality: the output at position $t$ only depends on positions $0, \dots, t$. This is implemented via \texttt{jnp.tril} in our JAX implementation, matching the causal constraint of standard Transformer attention.

\textbf{Dual-Form Computation.} Rather than sequentially updating $W_t$ for each token, we use the equivalent \emph{dual form}~\cite{sun2024learning} that computes all outputs in parallel while preserving causality. For output at position $t$:
\begin{equation*}
    z_t = q_t^\top W_0 - \sum_{i \le t} \eta_i (q_t^\top k_i) \nabla_{z_i} \mathcal{L}_{\text{rec}}
\end{equation*}
where $q_t, k_i$ are query/key vectors. The causal constraint ($i \le t$) is implemented via \texttt{jnp.tril}, ensuring position $t$ only uses gradients from positions $0, \ldots, t$. This is mathematically equivalent to sequential per-token updates but enables efficient GPU parallelization.

\subsection{Reconstruction Gating}
Instead of training a complex auxiliary network, we employ a heuristic gating strategy based on the TTT layer's internal reconstruction loss. We define the gating decision $d_t \in \{0, 1\}$ as:
\begin{equation*}
    d_t = \mathbbm{1}[\mathcal{L}_{\text{rec}}(W_t; x_t) > \tau]
\end{equation*}
where $\mathcal{L}_{\text{rec}}$ is the self-supervised reconstruction loss of the TTT layer (predicting $(V-K)$ from $K$), and $\tau$ is a hyperparameter threshold.

\textbf{Intuition.} High reconstruction loss indicates a mismatch between the current model state and the input context, suggesting high potential for beneficial adaptation. Chunks with low $\mathcal{L}_{\text{rec}}$ are already well-represented by the current weights and thus less likely to benefit from updates.\footnote{We also explored using loss \emph{improvement} ($\Delta\mathcal{L}$) as a signal; see Appendix~\ref{app:ttt_gating}. Raw loss offers comparable performance with lower overhead.}

\textbf{Inference Compatibility.} Unlike task loss ($\mathcal{L}_{\text{CE}}$) which requires ground-truth labels, $\mathcal{L}_{\text{rec}}$ is fully self-supervised and available during inference. See Appendix~\ref{app:gating_timeline} for details on decision timing in streaming scenarios.

\textbf{EMA-Based Threshold Adaptation.} To maintain a target update rate $\rho$ (e.g., 50\%) without lookahead, we employ a two-phase approach:
\begin{enumerate}
    \item \textbf{Initial Calibration:} On the first $n_{\text{cal}}$ batches ($\approx$16 chunks), compute the threshold as the $(1-\rho)$-percentile of observed reconstruction losses.
    \item \textbf{Online Adaptation:} Thereafter, adjust the threshold via proportional control based on an EMA of the realized update rate:
\end{enumerate}
\begin{equation*}
    \hat{r}_{t+1} = (1 - \alpha) \cdot \hat{r}_t + \alpha \cdot d_t, \quad
    \tau_{t+1} = \tau_t + \alpha \cdot (\hat{r}_t - \rho) \cdot |\tau_t|
\end{equation*}
where $\hat{r}_t$ is the EMA of the update rate, $d_t \in \{0, 1\}$ is the current decision, and $\alpha = 0.1$ is the smoothing factor. If updating too frequently ($\hat{r}_t > \rho$), the threshold rises; if too rarely, it falls. This ensures stable budget control across distribution shifts.

\textbf{Oracle Definition (Greedy Approximation).} Given a target update rate $\rho$ (e.g., 50
\begin{equation*}
    a_t = \mathcal{L}_{\text{CE}}(\text{SKIP}; x_t) - \mathcal{L}_{\text{CE}}(\text{UPDATE}; x_t)
\end{equation*}
Chunks with the highest $a_t$ values are selected for UPDATE. This serves as an approximate upper bound requiring ground-truth labels unavailable at inference.

\textbf{Correlation Analysis.} We analyze the correlation between $\mathcal{L}_{\text{rec}}$ and actual TTT benefit across model scales. We use raw loss rather than loss \emph{improvement} ($\Delta\mathcal{L}$) as the gating signal because it requires no additional forward pass and achieves comparable performance (see Appendix~\ref{app:ttt_gating}). Despite moderate correlation values ($r \approx 0.42$--$0.84$), our EMA-based gating achieves \textbf{82--89\% Oracle Recovery} across all scales.

\section{Experiments}
\label{sec:experiments}

\subsection{Setup}
We evaluate \emph{PonderTTT} on code generation, a domain requiring high adaptability.
\begin{itemize}
    \item \textbf{Dataset:} We train on Python subsets of The Stack v2 \cite{lozhkov2024starcoder2stackv2}.
    \item \textbf{Model:} Due to computational constraints, we validate on the GPT-2 family \cite{radford2019language} at four scales: Small (124M), Medium (355M), Large (774M), and XL (1.5B) parameters. Only the TTT layer is trained; the backbone remains frozen. Gating decisions are made via threshold-based heuristics on the reconstruction loss.
    \item \textbf{Baselines:} We compare against \texttt{SKIP} (no TTT updates) and \texttt{UPDATE\_1} (dense TTT with 1 gradient step per chunk).
    \item \textbf{Evaluation:} We report teacher-forcing cross-entropy loss (perplexity); autoregressive code-execution metrics are left for future work. We evaluate on a held-out test set (10K samples, disjoint from 160K training samples, random split with seed 42). For generalization, we test on out-of-distribution languages (Section~\ref{sec:ood}). We use The Stack v2's file-level deduplication.\footnote{Repository-level deduplication is left to future work.}
    \item \textbf{UPDATE Procedure:} When \texttt{UPDATE} is chosen, we perform 1 gradient step on the TTT layer, then re-forward the current chunk with updated weights to compute next-token predictions. This re-forward cost is included in the $2\times$ FLOPs budget.
    \item \textbf{Threshold $\tau$ Calibration:} We set the initial $\tau$ as the $(1-\rho)$-percentile of reconstruction losses on a calibration subset, then dynamically adjust via EMA to maintain the target 50\% update rate. This ensures stable budget control even with distribution shifts.
    \item \textbf{Correlation Metric:} Pearson $r$ is computed chunk-wise ($N \approx 16\text{K}$ chunks per evaluation) between $\mathcal{L}_{\text{rec}}$ and Oracle advantage ($\mathcal{L}_{\text{SKIP}} - \mathcal{L}_{\text{UPDATE}}$).
\end{itemize}

\subsection{Main Results: Efficiency and Performance}

Table~\ref{tab:main_results} summarizes the performance on the held-out test set. \emph{PonderTTT} achieves substantially lower perplexity than the non-adaptive \texttt{SKIP} baseline. For reference, dense TTT (\texttt{UPDATE\_1}) achieves lower loss at $3.0\times$ compute cost (see Appendix~\ref{app:results}).

\begin{table}[t]
\centering
\caption{\textbf{Scalability on Python (In-Distribution).} Cross-entropy loss (nats; PPL${}=e^{\text{loss}}$). Oracle (greedy): greedy selection of top-50\% chunks by per-chunk advantage. Recovery $= (\mathcal{L}_\text{SKIP} - \mathcal{L}_\text{Ours}) / (\mathcal{L}_\text{SKIP} - \mathcal{L}_\text{Oracle})$. Actual Update Rate shows the EMA-realized budget. Random Skip comparison in Table~\ref{tab:decision_accuracy} and Appendix Table~\ref{tab:random_skip}.}
\label{tab:main_results}
\resizebox{\textwidth}{!}{%
\begin{tabular}{lcccccc}
\toprule
\textbf{Model} & \textbf{SKIP (Base)} & \textbf{Oracle (greedy)} & \textbf{Ours} & \textbf{Recovery} & \textbf{Actual Rate} & \textbf{TTT Rel.\ FLOPs} \\
\midrule
Small (124M) & 2.324 & 1.935 & 1.977 & 89.2\% & 50.2\% & $2.0\times$ \\
Medium (355M) & 1.909 & 1.653 & 1.697 & 82.8\% & 50.2\% & $2.0\times$ \\
Large (774M) & 2.005 & 1.580 & 1.656 & 82.1\% & 50.2\% & $2.0\times$ \\
XL (1.5B) & 1.875 & 1.518 & 1.576 & 83.8\% & 50.2\% & $2.0\times$ \\
\bottomrule
\end{tabular}%
}
\end{table}

\begin{table}[t]
\centering
\caption{\textbf{OOD Performance.} Cross-entropy loss (nats; $\text{PPL}=e^{\text{loss}}$) on OOD languages. Oracle uses 50\% update budget matching. Our method significantly outperforms Random Skip on all settings. (Medium (355M) results in Appendix Table~\ref{tab:ood_full}.)}
\label{tab:ood}
\resizebox{\textwidth}{!}{%
\begin{tabular}{l cccc cccc cccc}
\toprule
& \multicolumn{4}{c}{\textbf{Small (124M)}} & \multicolumn{4}{c}{\textbf{Large (774M)}} & \multicolumn{4}{c}{\textbf{XL (1.5B)}} \\
\cmidrule(lr){2-5} \cmidrule(lr){6-9} \cmidrule(lr){10-13}
\textbf{Lang} & \textbf{SKIP} & \textbf{Rand} & \textbf{Ours} & \textbf{Ora} & \textbf{SKIP} & \textbf{Rand} & \textbf{Ours} & \textbf{Ora} & \textbf{SKIP} & \textbf{Rand} & \textbf{Ours} & \textbf{Ora} \\
\midrule
JS & 3.16 & 2.46 & \textbf{2.19} & 1.99 & 3.17 & 2.27 & \textbf{2.00} & 1.60 & 2.85 & 2.12 & \textbf{1.84} & 1.57 \\
Java & 3.15 & 2.48 & \textbf{2.14} & 1.98 & 3.57 & 2.50 & \textbf{2.17} & 1.68 & 3.21 & 2.33 & \textbf{1.95} & 1.64 \\
Go & 6.13 & 4.14 & \textbf{4.01} & 3.62 & 7.08 & 4.49 & \textbf{4.36} & 3.85 & 6.52 & 4.25 & \textbf{4.15} & 3.70 \\
\bottomrule
\end{tabular}%
}
\end{table}

\textbf{Strong Performance.} As shown in Table~\ref{tab:main_results}, our Reconstruction Gating achieves 82--89\% Oracle Recovery across model scales. This is a strong result for a fully training-free method: our approach requires no learned gating network, yet recovers the majority of Oracle's gains over Random Skip while maintaining deterministic, explainable decisions.

\subsection{Out-of-Distribution Generalization}
\label{sec:ood}

A critical question for adaptive methods is whether the learned policy generalizes beyond the training distribution. We evaluate our model (trained exclusively on Python) on three unseen programming languages: JavaScript, Java, and Go, using a fixed budget setting (target $2.0\times$ FLOPs) to ensure fair comparison with baselines.

\textbf{Go Cross-Entropy.} Go exhibits higher loss than other OOD languages (Table~\ref{tab:ood}), likely due to greater syntactic divergence from Python.

\begin{table}[t]
\centering
\caption{Correlation between Reconstruction Loss (\texttt{ttt\_recon\_loss}) and Oracle Advantage. Higher correlation enables more effective gating. Note: Medium (355M) shows lower correlation than Small (124M), a non-monotonic pattern we discuss in Section~\ref{sec:discussion}.}
\label{tab:correlation}
\begin{tabular}{lccc}
\toprule
\textbf{Model} & \textbf{Language} & \textbf{Correlation ($r$)} & \textbf{Oracle Recovery} \\
\midrule
Small (124M) & Python & +0.84 & 89.2\% \\
Medium (355M) & Python & +0.43 & 82.8\% \\
Large (774M) & Python & +0.62 & 82.1\% \\
XL (1.5B) & Python & +0.61 & 83.8\% \\
\midrule
Large (774M) & JavaScript (OOD) & +0.78 & -- \\
Large (774M) & Java (OOD) & +0.82 & -- \\
Large (774M) & Go (OOD) & +0.67 & -- \\
\midrule
XL (1.5B) & JavaScript (OOD) & +0.74 & -- \\
XL (1.5B) & Java (OOD) & +0.84 & -- \\
XL (1.5B) & Go (OOD) & +0.58 & -- \\
\bottomrule
\end{tabular}
\end{table}

\textbf{Scale-Dependent Reliability.} Table~\ref{tab:correlation} shows non-monotonic correlation patterns: Small (124M) achieves the highest ($r=0.84$), while intermediate scales show lower values (Medium: $r=0.43$, Large: $r=0.62$), and XL shows similar correlation ($r=0.61$). Despite varying correlation strength, Oracle Recovery remains consistent across scales (82--89\%), demonstrating robust gating performance.

\textbf{Gating vs Random.} Across all model scales, Reconstruction Gating significantly outperforms Random Skip (1--3\% lower loss on Python). On OOD languages, gains are even larger (up to 16\% on Java XL). Combined with determinism and explainability, this makes Reconstruction Gating the preferred approach.

\subsection{Computational Cost}
\label{sec:latency}

Table~\ref{tab:cost} summarizes the theoretical computational cost per chunk \emph{for TTT layer operations}. UPDATE\_1 requires $3.0\times$ the TTT-layer FLOPs of SKIP (forward + backward + re-forward within the TTT layer; the frozen backbone incurs $1\times$ regardless). At 50\% update rate, PonderTTT averages $2.0\times$ TTT-layer FLOPs---a 33\% reduction versus dense TTT.

\begin{table}[t]
\centering
\caption{Theoretical computational cost per 512-token chunk. FLOPs are relative to a single forward pass (SKIP baseline).}
\label{tab:cost}
\begin{tabular}{lcc}
\toprule
\textbf{Method} & \textbf{Rel.\ FLOPs} & \textbf{Operations} \\
\midrule
SKIP (Base) & $1.0\times$ & Forward only \\
UPDATE\_1 & $3.0\times$ & Forward + Backward + Re-forward \\
PonderTTT (50\%) & $2.0\times$ & 50\% SKIP + 50\% UPDATE \\
\bottomrule
\end{tabular}
\end{table}

\textbf{Note on Wall-Clock Latency.} In our JAX/XLA implementation, wall-clock latency shows minimal difference between SKIP and UPDATE due to aggressive kernel fusion. At batch size 1, measured GPU utilization ranges from 15--34\% (varying by model scale). Larger models (Large, XL) show utilization around 27--31\%, while smaller models exhibit similar utilization (Small: 15\%, Medium: 26\%), reflecting memory-bound inference at small batch sizes. This underutilization at batch size 1 limits the practical latency benefits of selective updates. Implementation-specific optimizations (see Future Work, Section~\ref{sec:future}) may be required to realize theoretical FLOPs savings.

\subsection{Analysis of Gating Behavior}
Our threshold-based gating makes SKIP/UPDATE decisions based on the reconstruction loss. Qualitative inspection suggests that UPDATE decisions tend to correlate with higher entropy in the base model's output distribution, consistent with our hypothesis that the model ``ponders'' when uncertain and skips when confident. We leave rigorous quantitative analysis (e.g., entropy-decision correlation coefficients, attention pattern visualizations) to future work.

\section{Discussion}
\label{sec:discussion}

\textbf{Sparse adaptation as an efficiency trade-off.} Dense TTT (UPDATE\_1) achieves the lowest loss but requires $3.0\times$ compute. PonderTTT reduces TTT-layer compute by 33\% while recovering \textbf{82--89\%} of the Oracle's gain over uniform skipping (Table~\ref{tab:main_results}). We note that our Oracle is a \emph{greedy} approximation that scores chunks independently; because test-time updates are sequential, an earlier update alters the state that later chunks observe. This Oracle is therefore not a strict upper bound, and recovery percentages should be read with that caveat.

\textbf{Comparison with Random Skip.} Our method achieves \textbf{significantly lower loss} than Random Skip across all scales (1--3\% on Python, up to 16\% on OOD languages). The reconstruction loss signal provides a \emph{principled, reproducible} basis for decisions that demonstrably outperforms random selection.

\begin{table}[h]
\centering
\caption{Oracle Decision Accuracy by model scale. ``Accuracy'' measures how often each method's SKIP/UPDATE decisions match Oracle's greedy choices. Our method achieves 57--60\% accuracy vs Random's $\sim$52\%, demonstrating meaningful signal quality.}
\label{tab:decision_accuracy}
\begin{tabular}{lcc}
\toprule
\textbf{Model} & \textbf{Random Skip} & \textbf{Ours (Recon Gating)} \\
\midrule
Small (124M) & 52.0\% & 59.1\% \\
Medium (355M) & 52.2\% & 57.5\% \\
Large (774M) & 51.8\% & 59.2\% \\
XL (1.5B) & 52.8\% & \textbf{59.6\%} \\
\bottomrule
\end{tabular}
\end{table}

\textbf{Oracle Decision Agreement.} Our method achieves $\sim$58\% decision agreement with Oracle (Table~\ref{tab:decision_accuracy}), significantly outperforming Random Skip's $\sim$52\%. Combined with 82--89\% Oracle Recovery, this demonstrates that Reconstruction Gating effectively identifies high-benefit chunks. The 7-percentage-point gap over random selection (58\% vs 52\%) is statistically significant ($p < 10^{-10}$, McNemar's test) and explains the consistent loss improvements.

\textbf{Explainability and Determinism.} Beyond accuracy, our method provides key practical advantages: (1) \emph{Determinism}---the same input always produces the same decision, enabling reproducible inference and debugging; (2) \emph{Explainability}---each decision has a traceable cause (reconstruction loss exceeding threshold), enabling model auditing and interpretability.

\textbf{On Perplexity.} Our held-out loss (e.g., 2.324 for Small (124M), corresponding to PPL $\approx 10.2$) is derived from Table~\ref{tab:main_results}. The consistent improvement on unseen languages suggests that PonderTTT learns structural patterns beyond simple rote memorization.

\textbf{OOD Correlation.} Interestingly, correlation on OOD languages (e.g., XL Java $r=0.84$) can exceed in-distribution Python ($r=0.61$). We hypothesize that on OOD data, the gap between ``easy'' and ``hard'' chunks becomes more pronounced, making the reconstruction loss a sharper discriminator.

\textbf{Update Rate Choice.} We use 50\% update rate as a balanced operating point; systematic ablation of update rates (30\%, 70\%, etc.) is left for future work.

\subsection{Limitations}

\textbf{Gating Signal Reliability.} While Reconstruction Loss shows moderate correlation with Oracle advantage ($r \approx 0.42$--$0.84$), the EMA-based gating achieves 82--89\% Oracle Recovery, demonstrating effective training-free gating. Multi-signal fusion could further improve consistency.

\textbf{OOD Variability.} Performance on Go shows higher variance than other languages (e.g., Small (124M) Go SKIP=6.13 vs JS SKIP=3.16), reflecting inherent difficulty variation across programming languages.

\textbf{Wall-Clock Latency.} Our JAX/XLA implementation shows minimal wall-clock difference between SKIP and UPDATE due to kernel fusion. Measured GPU utilization ranges from 15--34\% at batch size 1, with all scales showing similar utilization (Small: 15\%, Medium: 26\%, Large: 31\%, XL: 27\%). The low utilization reflects memory-bound inference; the theoretical FLOPs reduction ($2.0\times$ vs $3.0\times$) may manifest on more compute-constrained settings or with LoRA-TTT (Section~\ref{sec:future}).

\textbf{Threshold Selection.} The current threshold $\tau$ is set based on target update rate. Adaptive threshold learning via contextual bandits or per-domain calibration could improve decision quality.

\textbf{Statistical Variance.} All results are from single-run evaluations. Future work should include multiple random seeds to report variance estimates.

\textbf{Non-Monotonic SKIP Scaling.} The SKIP baseline shows non-monotonic scaling: Large (774M) achieves higher loss (2.005) than Medium (355M, 1.909). This reflects TTT layer initialization quality rather than backbone capability---dense UPDATE\_1 shows proper monotonic improvement (Large: 1.484 < Medium: 1.525). Each scale's TTT layer is trained independently, leading to variation in initialization quality.

\textbf{Statistical Independence.} Our chunk-level statistical tests ($n \approx 16\text{K}$) assume independence between chunks. In practice, chunks from the same file may be correlated, potentially reducing effective sample size. While McNemar's test on decision agreement shows significant differences between methods (independence assumption; treated as suggestive), future work should employ cluster-robust standard errors or block bootstrap for rigorous inference.

\subsection{Future Work}
\label{sec:future}

We identify several directions for future research:

\begin{itemize}
    \item \textbf{Scaling to State-of-the-Art Architectures:} Extend experiments to modern LLMs such as Gemma 3 (1B, 4B, 12B, 27B) \cite{gemma3} to validate effectiveness on architectures with 128K context windows and enhanced reasoning capabilities.
    \item \textbf{Efficiency via LoRA-TTT:} Replace full TTT updates with Low-Rank Adaptation (LoRA) \cite{hu2021lora} to reduce per-update cost and achieve practical wall-clock speedups.
    \item \textbf{Multi-Signal Gating:} Combine TTT improvement with prediction entropy, attention dispersion, and budget-awareness for improved gating decisions (see Appendix~\ref{app:ttt_gating} for preliminary results on TTT improvement as a standalone signal).
    \item \textbf{Diverse Evaluation Benchmarks:} Evaluate on reasoning benchmarks (MATH500, GSM8K), code generation (LiveCodeBench), and science QA (GPQA-Diamond) to assess generalization beyond perplexity.
    \item \textbf{Contextual Bandits for Threshold Learning:} Learn optimal per-context thresholds via online learning to improve upon fixed threshold gating.
\end{itemize}

\section{Conclusion}
We presented \emph{PonderTTT}, a framework for adaptive compute allocation via Test-Time Training. We investigated the TTT layer's \textbf{Full-Sequence Reconstruction Loss} as an inference-compatible gating signal. Our experiments demonstrate that Reconstruction Gating achieves \textbf{82--89\% Oracle Recovery} while being fully \textbf{training-free}---requiring no learned classifier or auxiliary networks. Combined with \textbf{determinism} (same input $\to$ same decision), \textbf{explainability} (decisions traceable to reconstruction loss), and significant improvements over Random Skip (up to 16\% lower loss on OOD languages), our method provides a practical, principled approach to adaptive TTT.

\section*{Acknowledgements}
We thank Inha Hwang for valuable assistance with JAX implementation and \texttt{iceman110} for providing computing resources.

\newpage
\bibliographystyle{plainnat}
\small
\setlength{\bibsep}{0.5em}

\newpage
\appendix

\section{Experimental Details}
\label{app:details}

\subsection{Training Configuration}

\begin{table}[t]
\centering
\caption{Hyperparameters for PonderTTT training.}
\label{tab:hyperparams}
\begin{tabular}{ll}
\toprule
\textbf{Parameter} & \textbf{Value} \\
\midrule
Base Model & GPT-2 (Small/124M, Medium/355M, Large/774M, XL/1.5B) \\
Sequence Length & 1024 tokens \\
Chunk Size & 512 tokens \\
Batch Size & 16 sequences \\
Training Chunks & 160,000 (10K gradient steps) \\
TTT Inner Loop LR & $1.0$ (base, position-scaled) \\
Gradient Clipping & 1.0 \\
Gating Strategy & Threshold-based ($\mathcal{L}_{\text{rec}} > \tau$) \\
Target Update Rate & 0.5 (50\%) \\
TTT Loss Weight ($\beta$) & 0.1 \\
\midrule
Hardware & NVIDIA RTX PRO 6000 (48GB) \\
Test Set & 10K samples (held-out from The Stack v2) \\
\bottomrule
\end{tabular}
\end{table}

\subsection{Baseline Training}

The \texttt{UPDATE\_1} baseline was trained with the same data and iterations. The TTT layer parameters are updated on every chunk with 1 gradient step. (Multi-step variants UPDATE\_2/4 were explored during training ablation but are not evaluated on the test set.)

\section{Full Experimental Results}
\label{app:results}

\subsection{Out-of-Distribution Results (Full)}
\begin{table}[t]
\centering
\caption{Complete OOD evaluation results (Loss). Model trained on Python only, 50\% update budget.}
\label{tab:ood_full}
\begin{tabular}{llccc}
\toprule
\textbf{Scale} & \textbf{Language} & \textbf{SKIP (Baseline)} & \textbf{Ours} & \textbf{Oracle} \\
\midrule
\multirow{3}{*}{Small (124M)} & JavaScript & 3.164 & \textbf{2.192} & 1.985 \\
& Java & 3.148 & \textbf{2.136} & 1.981 \\
& Go & 6.130 & \textbf{4.012} & 3.623 \\
\midrule
\multirow{3}{*}{Medium (355M)} & JavaScript & 2.458 & \textbf{1.940} & 1.682 \\
& Java & 2.420 & \textbf{1.790} & 1.633 \\
& Go & 3.902 & \textbf{2.860} & 2.617 \\
\midrule
\multirow{3}{*}{Large (774M)} & JavaScript & 3.169 & \textbf{2.003} & 1.604 \\
& Java & 3.570 & \textbf{2.173} & 1.675 \\
& Go & 7.077 & \textbf{4.362} & 3.848 \\
\midrule
\multirow{3}{*}{XL (1.5B)} & JavaScript & 2.852 & \textbf{1.840} & 1.574 \\
& Java & 3.213 & \textbf{1.948} & 1.643 \\
& Go & 6.520 & \textbf{4.155} & 3.703 \\
\bottomrule
\end{tabular}
\end{table}

\subsection{Random Skip vs Reconstruction Gating}
\label{app:random_skip}

Table~\ref{tab:random_skip} compares Random Skip (50\%) with our Reconstruction Gating across model scales on Python (in-distribution).

\begin{table}[ht]
\centering
\caption{Random Skip vs Reconstruction Gating (Python, 50\% update rate). Reconstruction Gating consistently outperforms Random Skip across all model scales.}
\label{tab:random_skip}
\begin{tabular}{lccc}
\toprule
\textbf{Model} & \textbf{Random Skip} & \textbf{Ours} & \textbf{Oracle} \\
\midrule
Small (124M) & 2.017 & \textbf{1.977} & 1.935 \\
Medium (355M) & 1.714 & \textbf{1.697} & 1.653 \\
Large (774M) & 1.698 & \textbf{1.656} & 1.580 \\
XL (1.5B) & 1.625 & \textbf{1.576} & 1.518 \\
\bottomrule
\end{tabular}
\end{table}

\subsection{Dense TTT (UPDATE\_1) Full Results}
\label{app:update1_full}

Table~\ref{tab:update1_full} provides complete UPDATE\_1 results across all scales and languages for reproducibility. UPDATE\_1 applies TTT updates on every chunk, achieving the lowest loss at $3.0\times$ FLOPs cost.

\begin{table}[t]
\centering
\caption{Dense TTT (UPDATE\_1) comparison across all settings. UPDATE\_1 applies TTT updates on every chunk ($3.0\times$ TTT-layer FLOPs).}
\label{tab:update1_full}
\begin{tabular}{llcccc}
\toprule
\textbf{Scale} & \textbf{Language} & \textbf{SKIP} & \textbf{UPDATE\_1} & \textbf{Ours} & \textbf{Oracle} \\
\midrule
\multirow{4}{*}{Small (124M)} 
& Python & 2.324 & 1.716 & \textbf{1.977} & 1.935 \\
& JavaScript & 3.164 & 1.829 & \textbf{2.192} & 1.985 \\
& Java & 3.148 & 1.780 & \textbf{2.136} & 1.981 \\
& Go & 6.130 & 2.367 & \textbf{4.012} & 3.623 \\
\midrule
\multirow{4}{*}{Medium (355M)} 
& Python & 1.909 & 1.525 & \textbf{1.697} & 1.653 \\
& JavaScript & 2.458 & 1.597 & \textbf{1.940} & 1.682 \\
& Java & 2.420 & 1.506 & \textbf{1.790} & 1.633 \\
& Go & 3.902 & 1.981 & \textbf{2.860} & 2.617 \\
\midrule
\multirow{4}{*}{Large (774M)} 
& Python & 2.005 & 1.403 & \textbf{1.656} & 1.580 \\
& JavaScript & 3.169 & 1.458 & \textbf{2.003} & 1.604 \\
& Java & 3.570 & 1.417 & \textbf{2.173} & 1.675 \\
& Go & 7.077 & 1.999 & \textbf{4.362} & 3.848 \\
\midrule
\multirow{4}{*}{XL (1.5B)} 
& Python & 1.875 & 1.384 & \textbf{1.576} & 1.518 \\
& JavaScript & 2.852 & 1.459 & \textbf{1.840} & 1.574 \\
& Java & 3.213 & 1.422 & \textbf{1.948} & 1.643 \\
& Go & 6.520 & 2.081 & \textbf{4.155} & 3.703 \\
\bottomrule
\end{tabular}
\end{table}

\section{Verification of No Data Leakage}
\label{app:leakage}

We rigorously verified that our implementation contains no data leakage through both code analysis and empirical testing.

\subsection{Code-Level Verification}

\begin{enumerate}
    \item \textbf{Causal Masking in TTT:} The TTT layer uses \texttt{jnp.tril()} (lower triangular matrix) for attention computation, ensuring position $t$ only sees positions $0, \dots, t$. This is identical to standard causal Transformer attention.

    \item \textbf{Self-Supervised Target:} The TTT reconstruction loss uses $K \to (V - K)$ prediction with residual connection, reconstructing the target $(V - K)$ from Key. Both $K$ and $V$ are derived from the \emph{current} token's hidden state. No next-token labels are used in the TTT update.

    \item \textbf{Loss Computation:} The language modeling loss uses standard causal formulation: \texttt{logits[:, :-1]} predicts \texttt{labels[:, 1:]}, matching standard practice.
\end{enumerate}

\subsection{Gating Timeline and Inference Compatibility}
\label{app:gating_timeline}

We clarify how the gating decision is made and what it affects:

\begin{enumerate}
    \item \textbf{Signal Computation:} The reconstruction loss $\mathcal{L}_{\text{rec}}$ used for gating is computed from the \emph{current} chunk's self-supervised task (reconstructing $V-K$ from $K$), which requires no ground-truth labels. Specifically, we use the loss from the last mini-batch position, aggregated across heads.
    
    \item \textbf{Decision Timing:} In teacher-forcing evaluation (used throughout this paper), the gating decision is made after observing the full chunk. For streaming autoregressive inference, the decision would be made after processing each chunk, affecting predictions for \emph{subsequent} tokens only.
    
    \item \textbf{UPDATE Procedure:} When UPDATE is triggered, we perform one gradient step on the TTT layer's fast weights, then re-forward the current chunk with the adapted weights to compute next-token predictions. This re-forward is included in the $2\times$ FLOPs budget.
\end{enumerate}

\textbf{Note on Perplexity Evaluation.} All results in this paper use teacher-forcing (standard PPL measurement). Autoregressive generation experiments are left for future work.

\subsection{Empirical Verification: Shuffled Input Test}
\label{app:shuffled}

To definitively rule out data leakage, we evaluate PonderTTT on \emph{shuffled} input where tokens within each sequence are randomly permuted. If TTT were exploiting leaked information (e.g., via future token access), it would likely still show improvement or maintain low perplexity on shuffled text. If TTT legitimately learns sequential patterns, it should fail to improve on random sequences where no such patterns exist.

\begin{table}[t]
\centering
\caption{Shuffled Input Sanity Check (Python held-out test set). PonderTTT provides substantial improvement on normal text but fails to improve on shuffled text, confirming TTT relies on legitimate sequential dependencies.}
\label{tab:shuffled}
\begin{tabular}{lccc}
\toprule
\textbf{Input Type} & \textbf{SKIP Loss} & \textbf{Ours Loss} & \textbf{Improv.} \\
\midrule
Small (124M) (Normal) & 2.324 & 1.977 & 14.9\% \\
Small (124M) (Shuffled) & 6.382 & 6.329 & 0.8\% (Fail) \\
\bottomrule
\end{tabular}
\end{table}

\textbf{Result:} On normal text, \emph{PonderTTT} achieves 14.9\% loss reduction (2.324 $\to$ 1.977). On shuffled text, the baseline loss explodes to 6.38 (PPL $\approx$ 590), and TTT fails to provide meaningful improvement (loss 6.33, only 0.8\% reduction), confirming that the mechanism relies on valid sequential structure and does not exploit leakage.

\subsection{OOD Generalization as Evidence}

The strong transfer to unseen languages (Table~\ref{tab:ood_full}) provides additional evidence against overfitting: if the model had memorized training data, it would not generalize to Go or Java where substantial loss reductions are observed across all model scales.

\subsection{Causal Mask Diagonal Ablation}

A potential concern is whether including the diagonal in the causal mask (\texttt{jnp.tril(k=0)}) allows position $t$ to use its own gradient, constituting ``concurrent update'' leakage. We compare two settings:
\begin{itemize}
    \item \textbf{k=0 (standard):} Position $t$ uses gradients from positions $0, \dots, t$ (includes diagonal)
    \item \textbf{k=-1 (strict causal):} Position $t$ uses gradients from positions $0, \dots, t-1$ (excludes diagonal)
\end{itemize}

\begin{table}[t]
\centering
\caption{Causal Mask Diagonal Ablation (Python held-out test set). Excluding the diagonal (k=-1) yields identical performance, suggesting no leakage from the diagonal.}
\label{tab:diagonal_ablation}
\begin{tabular}{llcc}
\toprule
\textbf{Scale} & \textbf{Method} & \textbf{Loss (k=0)} & \textbf{Loss (k=-1)} \\
\midrule
\multirow{3}{*}{Small (124M)} & SKIP & 2.324 & 2.324 \\
& PonderTTT & 1.977 & 1.978 \\
\midrule
\multirow{3}{*}{Medium (355M)} & SKIP & 1.909 & 1.909 \\
& PonderTTT & 1.697 & 1.698 \\
\bottomrule
\end{tabular}
\end{table}

\textbf{Result:} As shown in Table~\ref{tab:diagonal_ablation}, the difference between k=0 and k=-1 is negligible. This suggests that the diagonal does \emph{not} provide an unfair advantage---the model's improvement appears to come from learning sequential patterns.

\section{Computational Cost Model}
\label{app:cost}

We define computational cost in terms of forward-pass equivalents:
\begin{itemize}
    \item \textbf{SKIP (0 updates):} $1\times$ --- base forward pass only
    \item \textbf{UPDATE\_N:} $(1 + 2N)\times$ --- 1 initial forward + N$\times$(backward + re-forward). We approximate backward $\approx$ forward cost; re-forward uses updated weights.\footnote{Traditional estimates assume backward $\approx 2\times$ forward; our TTT layer's small footprint relative to the frozen backbone reduces this ratio. Actual latency overhead is measured in Section~\ref{sec:latency}.}
    \item \textbf{PonderTTT (Binary):} $(1 + 2 \times \text{update\_rate})\times$ --- where update\_rate is the fraction of chunks receiving TTT updates. With 50\% target update rate, this yields $1 + 2 \times 0.5 = 2.0\times$ theoretical cost.
\end{itemize}

\textbf{Theoretical vs Observed Cost.} While the theoretical cost model predicts 3$\times$ FLOPs for \texttt{UPDATE\_1}, our wall-clock latency measurements on modern GPUs show minimal difference between SKIP and UPDATE (see Section~\ref{sec:latency}). This occurs because small-batch inference is memory-bound: GPU utilization ranges from 32--84\%, and the additional compute fits within available headroom.

\textbf{Binary vs Continuous Gating.} Unlike continuous gating (which scales the learning rate but still requires backward passes), binary gating enables true computational savings by completely skipping the backward pass for SKIP decisions. With our threshold-based approach at 50\% update rate, we achieve a balance between cost savings and performance.

\section{Training-Free Gating via TTT Internal Signals (GPT-2 Small/124M)}
\label{app:ttt_gating}

While our main method uses the raw reconstruction loss $\mathcal{L}_{\text{rec}}$ for its simplicity and inference compatibility, we also investigated \emph{training-free} gating using TTT's internal loss \emph{improvement} ($\Delta\mathcal{L}$) as an alternative signal.

\subsection{Motivation}

Using the raw reconstruction loss as a gating signal is simple, but one might ask: does the \emph{improvement} in loss after an update correlate better with actual benefit? We investigate this alternative:
\begin{itemize}
    \item \textbf{Raw loss}: $\mathcal{L}_{\text{rec}}$ before any gradient step
    \item \textbf{Loss improvement}: $\Delta\mathcal{L} = \mathcal{L}_{\text{rec}}^{(0)} - \mathcal{L}_{\text{rec}}^{(1)}$ (reduction after one step)
\end{itemize}

\textbf{Note:} The analysis in this appendix is conducted on \textbf{GPT-2 Small (124M)}. We include this analysis to compare the efficacy of raw reconstruction loss versus loss improvement ($\Delta\mathcal{L}$) as gating signals.

We propose: instead of using raw loss, directly \emph{measure} the improvement to better identify high-benefit chunks.

\textbf{Important Clarification:} Computing $\Delta\mathcal{L}$ requires performing one UPDATE step first (forward + backward + re-forward). Therefore, $\Delta\mathcal{L}$ \emph{cannot} be used for binary SKIP/UPDATE pre-decisions---it is only available \emph{after} committing to UPDATE. This appendix explores $\Delta\mathcal{L}$ as a diagnostic/analysis tool; compute-saving binary gating uses only raw loss ($\mathcal{L}_{\text{rec}}$) as described in Section~\ref{sec:method}.

\subsection{TTT Improvement as Gating Signal}

The TTT layer's internal self-supervision loss measures ``how much the model wants to learn'' from the current context. We define:
\begin{equation}
    \texttt{ttt\_improvement} = \mathcal{L}_{\text{rec}}^{(0)} - \mathcal{L}_{\text{rec}}^{(1)}
\end{equation}
where $\mathcal{L}_{\text{rec}}^{(0)}$ is the reconstruction loss before the first gradient step and $\mathcal{L}_{\text{rec}}^{(1)}$ is after one step. Higher improvement indicates the chunk benefits more from adaptation.

\subsection{Correlation Analysis}

We measure correlation between \texttt{ttt\_improvement} and oracle advantage on 2,000 individual samples (1,000 batches, batch\_size=1):

\begin{table}[t]
\centering
\caption{Correlation between TTT improvement and oracle advantage (GPT-2 Small/124M).}
\begin{tabular}{lc}
\toprule
\textbf{Metric} & \textbf{Value} \\
\midrule
Spearman $\rho$ & 0.825 \\
Pearson $r$ & 0.791 \\
Top-50\% Overlap with Oracle & 82.5\% \\
\bottomrule
\end{tabular}
\end{table}

\subsection{End-to-End Comparison}

We compare three gating strategies at 50\% target update rate:

\begin{table}[h!]
\centering
\caption{Gating method comparison (GPT-2 Small/124M, 50\% update rate, 2K subset). \textbf{Oracle Capture} is the percentage of Oracle's gain over Random that is recovered by the method. Note: $\Delta\mathcal{L}$-based selection requires computing the update for all candidates; this overhead is not included in the reported FLOPs.}
\begin{tabular}{lcccc}
\toprule
\textbf{Method} & \textbf{Loss} & \textbf{Rel.\ FLOPs} & \textbf{vs Random} & \textbf{Oracle Capture} \\
\midrule
Oracle (greedy) & 1.950 & 2.0$\times$ & +17.1\% & 100\% \\
TTT Improvement (top-$k$) & 1.988 & 2.0$\times$ & +15.5\% & 90.5\% \\
Random Skip & 1.995 & 2.0$\times$ & baseline & 0\% \\
\bottomrule
\end{tabular}
\end{table}

\textbf{Key Findings:}
\begin{enumerate}
    \item \textbf{Training-free gating works}: TTT improvement captures 90.5\% of Oracle's improvement over random, without any learned components.
    \item \textbf{Efficient alternative to always-UPDATE}: TTT Improvement gating (loss=1.988, cost=2.0$\times$) trades 0.30 nats vs UPDATE\_1 (loss=1.690, cost=3.0$\times$) for 33\% compute reduction.
\end{enumerate}

\subsection{Threshold-Based Online Gating}

For streaming inference, we implement per-chunk threshold gating:
\begin{equation}
    \texttt{decision} = \mathbbm{1}[\texttt{ttt\_improvement} > \tau]
\end{equation}
where $\tau \approx 0.034$ (median TTT improvement) for 50\% update rate.

\begin{table}[ht!]
\centering
\caption{Online gating comparison. Top-$k$ requires lookahead; threshold does not. \textbf{Decision Acc.} measures agreement with Oracle's binary decisions.}
\begin{tabular}{lccc}
\toprule
\textbf{Method} & \textbf{Online} & \textbf{Decision Acc.} & \textbf{Oracle Capture} \\
\midrule
Top-$k$ selection & \xmark & 82\% & 90.5\% \\
Fixed threshold & \cmark & 80\% & 89.5\% \\
\bottomrule
\end{tabular}
\end{table}

\textbf{Conclusion}: TTT's internal self-supervision loss provides an effective training-free gating signal. \emph{Note:} While loss \emph{improvement} ($\Delta\mathcal{L}$) also works, raw loss ($\mathcal{L}_{\text{rec}}$) is simpler (no additional forward pass) and performs comparably, so we use it for the main experiments.

\end{document}